\documentclass[sigconf,screen]{acmart}

\usepackage{lscape}
\usepackage{dsfont}
\usepackage{multirow}
\usepackage{adjustbox}
\AtBeginDocument{%
  \providecommand\BibTeX{{%
    \normalfont B\kern-0.5em{\scshape i\kern-0.25em b}\kern-0.8em\TeX}}}

\copyrightyear{2024}
\acmYear{2024}
\setcopyright{acmlicensed}\acmConference[AIMLSystems 2024]{The Fourth International Conference on Artificial Intelligence and Machine Learning Systems}{}{Louisiana State University, Baton Rouge, LA}
\acmBooktitle{The Fourth International Conference on Artificial Intelligence and Machine Learning Systems (AIMLSystems 2024), October 25--28, 2024, Bangalore, India}
\acmDOI{}
\acmISBN{979-8-4007-1649-2/23/10}

\DeclareMathOperator{\argmax}{argmax}

\begin{document}

\title{Assessing the Impact of Upselling in Online Fantasy Sports}

\author{Aayush Chaudhary}
\affiliation{%
  \institution{Dream11}
  \city{Mumbai}
  \country{India}
}
\email{aayush@dream11.com}

\renewcommand{\shortauthors}{Chaudhary, et al.}

\begin{abstract}
This study explores the impact of upselling on user engagement. , we model users' deposit behaviour on the fantasy sports platform Dream11. Subsequently, we develop an experimental framework to evaluate the effect of upselling using an intensity parameter. Our live experiments on user deposit behaviour reveal decreased user recall with heightened upselling intensity. Our findings indicate that increased upselling intensity improves user deposit metrics and concurrently diminishes user satisfaction and conversion rates. We conduct robust counterfactual analysis and train causal meta-learners to personalise user's upselling intensity levels to reach an optimal trade-off point.

\end{abstract}


\maketitle

\section{Introduction}
Upselling or cross-selling refers to offering users additional products or services that are higher-end or complementary to their initial purchase. This strategy is widely employed across various industries to enhance revenue and customer value. According to Schiffman \cite{schiffman2005upselling}, "Upselling is the initiative to encourage customers who have already purchased to buy more of the same product or additional products.", whereas Kubiak et al. (2010) describe upselling as advancing to a more expensive product or service \cite{kubiak2010upselling}.

Upselling is studied extensively across multiple sectors, such as airlines, hospitality, retail, and insurance. And their impacts on customer satisfaction and revenue metrics. For instance, in the airline industry, upselling involves flight seat upgrades \cite{thirumuruganathan2023will}, while in the hospitality industry, it involves luxury room upgrades \cite{guillet2020online}, In the Insurance industry \cite{Guelman2014ASO} to cross-sell/Upsell. 

Upselling applications span sectors. In this paper, we study user behaviour of upselling on a fantasy sports platform, Dream11, with over 200M users. Further, to optimise the upselling policy, we developed a system that has not been effectively addressed by existing methods relying on traditional machine learning techniques.

In this work, we draw the following \textit{conclusions} to the field of user behaviour modelling in upselling on online platforms:
\begin{itemize}
    \item Using a multi-class supervised model, we predict the deposit propensity of a user and evaluate it across supervised regression, classification and heuristic baseline methods. During online experiments, we analysed the trade-offs between recall and revenue uplift. 
    
    \item Evaluate causal models to address the trade-offs and arrive at an optimal upselling policy from \textit{real-world deposit data} collected using grid experiments across the spectrum of intensity values.\end{itemize}

By accurately predicting deposit amounts and tailoring upselling suggestions, we aim to ensure a seamless payment system, user engagement, fairness, and the platform's revenue optimisation. We conducted several online grid experiments across intensity gradients to upsell users. We found that increasing intensity leads to a decrease in user recall. The recall dropped because some users disliked upselling, and many new users had dropped off the conversion funnel. Therefore, it was evident that there should be a personalised upselling policy. The proposed approach builds upon previous work in uplift modelling to estimate heterogeneous treatment effects \cite{Zhao2019UpliftMF}.

\section{Problem Formulation and Algorithms}
We develop a supervised model to predict the expected deposit amount for each user. This model will feed the upselling strategy by generating a list of suggested deposit amounts tailored to individual users.

\subsection{Supervised Model Formulation}
Let \( \mathbf{X} = \{x_1, x_2, \ldots, x_n\} \) denote the set of user-specific and deposit-specific features, including demographic information, historical transaction data, and engagement metrics. 
Let \(F(\mathbf{X})\) represent the set of deposit amounts for a user, modelled as a multi-class classification model. The problem involves two key components: predicting the expected deposit amount for each user and computing the recommended list of suggested deposit amounts. We want to learn the function \( F(\mathbf{X}) \) from the historical deposit data \( \mathbf{X} \) to predict the expected deposit amounts. 

\subsubsection{Evaluation and Learnings}
In this paper, we introduce key metrics designed to provide insights into the accuracy and reliability of our classification models. Since our dataset was imbalanced towards lower deposit amounts, we selected the weighted F1 score as our evaluation criterion because it is crucial to have a model that is equally correct for all the classes. The dataset is imbalanced with lower-value transaction amounts, causing a significant load on the payment systems. Focal loss and l2 regularization proved pivotal for the accuracy of the final model. 

\begin{table}[h]
\centering
\begin{tabular}{cc}
\textbf{Model} & \textbf{F1 Score (Weighted)} \\
\hline
Heuristic (Median rolling 10 Txn) & 0.736 \\
LGBM Regressor & 0.758 \\
LGBM Classifier & 0.804 \\
LGBM Classifier (Focal Loss) & 0.852 \\
\end{tabular}
\caption{Model Performance Comparison}
\end{table}

\section{Experimentation}
\subsection{Upselling Experiment}
The payments dataset used in this research comes from millions of payment transactions conducted on the Dream11 fantasy sports gaming platform. Whenever the user adds cash to the account, we recommend a set of three values shown in table \ref{tab:recommendations}. There are maximum limits and UX constraints on these values. We arrived at these values by analyzing historical deposit amounts and identifying the most frequent deposit amounts and their multiples. This allowed us to cover a range of common deposit behaviours and encourage higher deposits without overwhelming the user.

\begin{table}[h]
\centering
\begin{tabular}{|c|c|c|}
\hline
\multicolumn{3}{|c|}{Amount to add} \\
\hline
Prefill value & Option 1 & Option 2 \\
\hline
50 & 200 & 500 \\
\hline
\end{tabular}
\caption{Recommended Deposit Values}
\label{tab:recommendations}
\end{table}

\subsubsection{Objective:} The aim was to evaluate the impact of upselling using an intensity factor (\(a\)). We sampled deposit amounts from the user-specific distribution \( \text{F}(X_{i}) \) and adjusted them using \(a\). Users were randomly assigned to one of the four target groups (TGs) listed in Table \ref{tab:target-groups}.

        \begin{table}[h]
        \centering
        \begin{tabular}{cc}
        \textbf{Target Group} & \textbf{Treatment} \\
        \hline
        TG1 & 1x \\
        TG2 & 1.25x \\
        TG3 & 1.5x \\
        TG4 & 2x \\
        \end{tabular}
        \caption{Intensity Treatment Groups}
        \label{tab:target-groups}
        \end{table}
        
\subsubsection{Guardrails}
We monitored the following metrics to ensure upselling didn't negatively affect user experience:
\begin{itemize}
    \item \textbf{Conversion Funnel:} Ratio of successful transactions to attempted transactions.
    \item \textbf{User Engagement:} Recall on the recommendations, Time spent on the platform and changes in transaction frequency.
    \item \textbf{Revenue Uplift:} Comparison of average deposit amounts across target groups.
\end{itemize}

\subsection{Online Results}
\begin{figure}[htbp]
    \centering
    \includegraphics[width=1\linewidth]{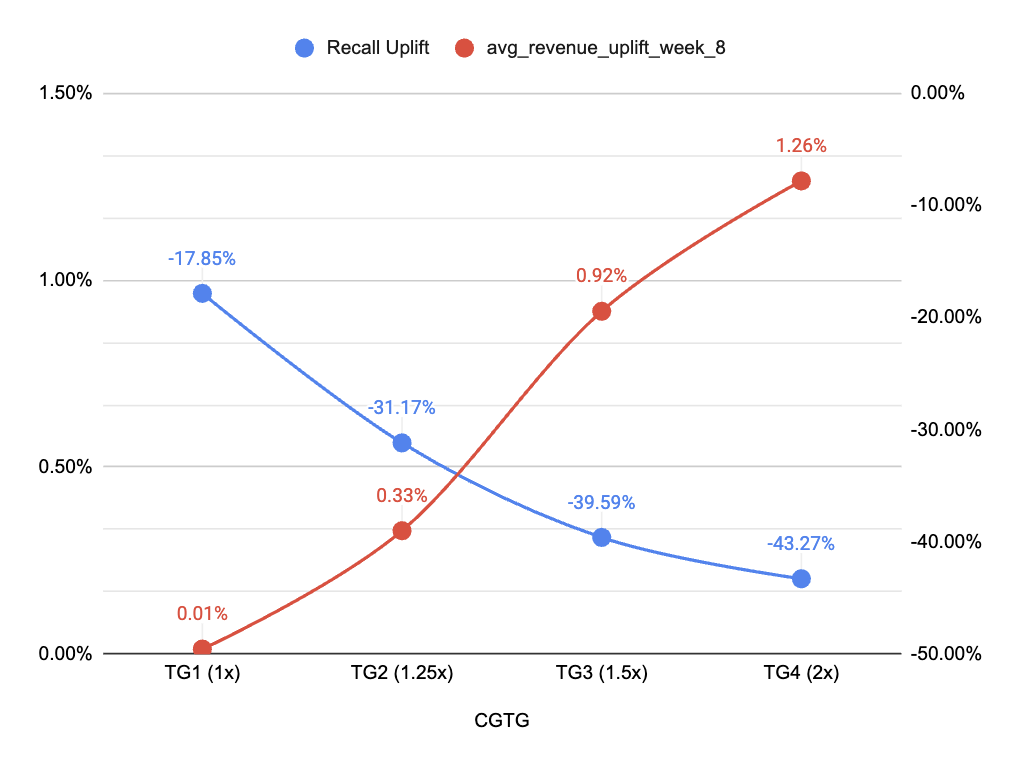}
    \caption{Recall and Revenue Uplift Across Variants}
    \label{fig:sensitivity}
\end{figure}

The control group received a default recommendation, whereas the target groups received various personalized amounts alongside different discrete intensity values [1x, 1.25x, 1.5x, 2x]. This experiment lasted eight weeks, from January 1, 2024, to February 28, 2024.
We observed an immediate impact on the per transaction value of about +5.6\%, while users were also making fewer transactions on the platform (reduction in infra cost), resulting in a 4.8\% fall in the most aggressive variant. While evaluating the long-term sustained impact in the avg\_deposit\_week\_8, Figure [\ref{fig:sensitivity}] illustrates the uplift across the experimental variants shown in Table [\ref{tab:target-groups}]. 

Upselling impacted the experiment guardrails, which were the user engagement metrics. The time spent for a successful payment increased, and user conversion dropped by almost 2.1\%, mainly impacting new users.

\section{Causal Uplift Modelling}
During data analysis of the upselling experiment, we saw that the challenge was to estimate how much to upsell a user without a trade-off in user satisfaction. A popular approach for efficiently assigning treatments to a subset of customers for targeted personalization is using causal machine learning models \cite{Gutierrez2017CausalIA, Diemert2018ALS}. 

\subsection{Dataset Preparation}
For causal uplift modeling, the dataset consists of three key components: Treatment (T), Response (Y), and Features/Covariates (X). These are defined as follows:

\begin{itemize}
    \item \textbf{Treatment (T\textsubscript{i}):} The intensity of upselling, represented by four levels: \{1.0, 1.25, 1.50, 2.0\}.
    \item \textbf{Response (Y):} The primary outcome of interest, measured as the total deposit amount by each user.
    \item \textbf{Features (X\textsubscript{i}):} User-specific characteristics, including demographics, gameplay behavior, and historical transaction data.
\end{itemize}

To arrive at an optimal upselling policy, we personalize the intensity factor for each user. We estimate the incremental uplift by applying the intensity treatments (T\textsubscript{i}) uniformly across all users. This dataset will then be used to model and optimize upselling strategies.

\subsection{Model Training}
We estimated the Conditional Average Treatment Effect (CATE) \( Y(x) \), which represents the change in the probability of completing a deposit \( \Pr(Y = 1) \) due to varying upselling intensity, based on user features \(X\). To achieve this, we tested four algorithms: S-learner, T-learner, X-learner, and R-learner, implemented using Python’s \texttt{causalml} library \cite{chen2020causalml}, with XGBRegressor as the base model and a fixed propensity score of 0.5.

For model training and optimization, we used the \texttt{hyperopt} package \cite{Bergstra2013HyperoptAP} with SparkTrials to distribute workloads, and MLflow \cite{Zaharia2018AcceleratingTM} to manage model runs. The parameter search was guided by Tree-based Parzen Estimators (TPE), with the objective function defined as:

\begin{itemize}
    \item Define the parameter search space using \texttt{causalml}.
    \item Train models on the training dataset.
    \item Apply trained models to both training and test sets.
    \item Optimize by calculating the mean AUUC (Area Under the Uplift Curve) score on the test set.
\end{itemize}

We ran 1000 random trials, logging the AUUC for each treatment. Two model configurations were tested: 
\begin{itemize}
    \item \textbf{Global configuration}: Selects the model with the lowest test set loss across all treatments.
    \item \textbf{Local configuration}: Optimizes the CATE for each treatment group (TG) using the highest test set AUUC.
\end{itemize}

Table \ref{table:xgb_configs} summarizes the XGBRegressor configurations used in the SparkTrials. Each treatment group (TG1, TG2, TG3, TG4) has its own local configuration parameters, while the global model (TG\_global) is optimized for overall performance. A treatment is assigned based on the following rule:

\begin{equation*}
\pi(x) = 
\begin{cases} 
\argmax\limits_{t \neq \text{CG}} \text{CATE}(x \mid T = t) & \text{if} \; \max\limits_{t \neq \text{CG}} \text{CATE}(x \mid T = t) > 0 \\
\text{CG} & \text{otherwise}
\end{cases}
\end{equation*}

\begin{table}[h]
\centering
\begin{adjustbox}{max width=\linewidth}
\begin{tabular}{lccccc}
\toprule
\textbf{Parameter} & \textbf{TG1} & \textbf{TG2} & \textbf{TG3} & \textbf{TG4} & \textbf{TG\_global} \\
\midrule
meta\_learner & r\_learner & x\_learner & r\_learner & s\_learner & r\_learner \\
gamma & 1.24 & 8.40 & 7.95 & 4.60 & 5.67 \\
n\_estimators & 59 & 153 & 102 & 125 & 200 \\
colsample\_bytree & 0.82 & 0.56 & 0.93 & 0.88 & 0.75 \\
max\_depth & 7 & 10 & 15 & 8 & 12 \\
min\_child\_weight & 5 & 3 & 6 & 4 & 2 \\
reg\_lambda & 0.36 & 0.14 & 0.57 & 0.94 & 0.50 \\
reg\_alpha & 77 & 115 & 50 & 163 & 90 \\
\midrule
\textbf{Configuration} & \multicolumn{4}{c}{\textbf{Local config}} & \textbf{Global config} \\
\bottomrule
\end{tabular}
\end{adjustbox}
\caption{XGBRegressor Configurations for Each Treatment Group}
\label{table:xgb_configs}
\end{table}

\subsection{Evaluation}
Uplift modelling faces performance limitations due to the need for an unbiased evaluation metric. Individual responses to actions and natural responses are unknown simultaneously, preventing explicit labelling of uplift responses. We underwent rigorous evaluations on both the business and model sides to establish confidence in our analysis.
\begin{itemize}
    \item \textbf{Percent Treated:} Percentage of users distributed across all possible treatment values, including control. In Table [\ref{table:policy_assignment}], the grid-experiment policy is uniformly distributed while the CATE policy is assigned with a non-negative threshold.  

\begin{table}[h]
\centering
\begin{tabular}{ccc}
\toprule
\textbf{Percent Treated} & \multicolumn{2}{c}{\textbf{policy}} \\
\cmidrule(lr){2-3}
\textbf{Treatment} & \textbf{Grid exp policy} & \textbf{CATE policy($\pi$)} \\
\midrule
CG & 19.92 & 3.16 \\
TG1 & 20.12 & 12.71 \\
TG2 & 20.05 & 31.86 \\
TG3 & 20.13 & 37.42 \\
TG4 & 19.78 & 14.86 \\
\bottomrule
\end{tabular}
\caption{Policy Assignment Table}
\label{table:policy_assignment}
\end{table}

 \item \textbf{ERUPT:} Expected Response Under Proposed Treatments: \cite{Zhao2017UpliftMW}, \cite{Hitsch} It quantifies a model's performance at the downstream task of predicting the optimal treatment to assign each member of the population. It is defined as follows 
 \begin{equation}
\text{ERUPT} = \mathbb{E}(Y \cdot \mathbb{I}(\pi(x) = t) \mid T = t, \pi) \end{equation}

We calculate the expected bootstrapped revenue using a sample size of 5000. After model inference, the CATE policies for both test (\(\pi_{\text{test}}\)) and train (\(\pi_{\text{train}}\)) datasets are derived, with the intensity factor yielding the highest uplift being selected. As shown in Figure [\ref{fig:erupt}], the CATE policy results in a significant 10.7\% uplift across both the training and test datasets.
\newline
\begin{figure}
    \centering
    \includegraphics[width=1\linewidth]{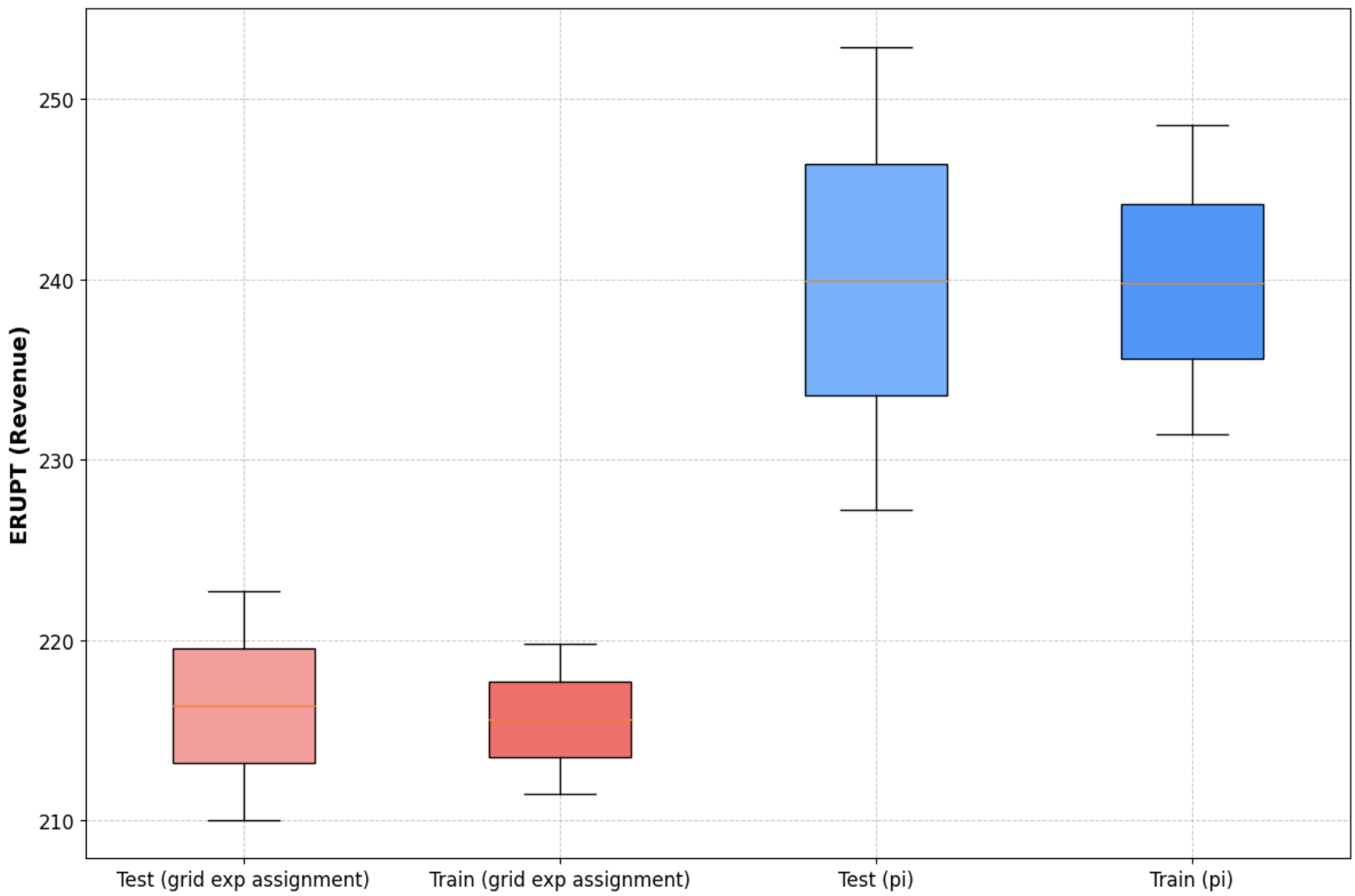}
    \caption{Erupt revenue distribution for grid experiment policy vs CATE policy across test and train datasets.}
    \label{fig:erupt}
\end{figure}
\item \textbf{AUUC (Area Under the Uplift Curve):} 
The uplift chart consists of an Uplift Curve and a random baseline curve for each treatment. The charts demonstrate a strong offline fit of the underlying estimators. The cumulative gains are plotted by ranking users based on their inferred uplift scores, showing that most gains are concentrated within the top-ranked users.

\begin{figure}
    \centering
    \includegraphics[width=1\linewidth]{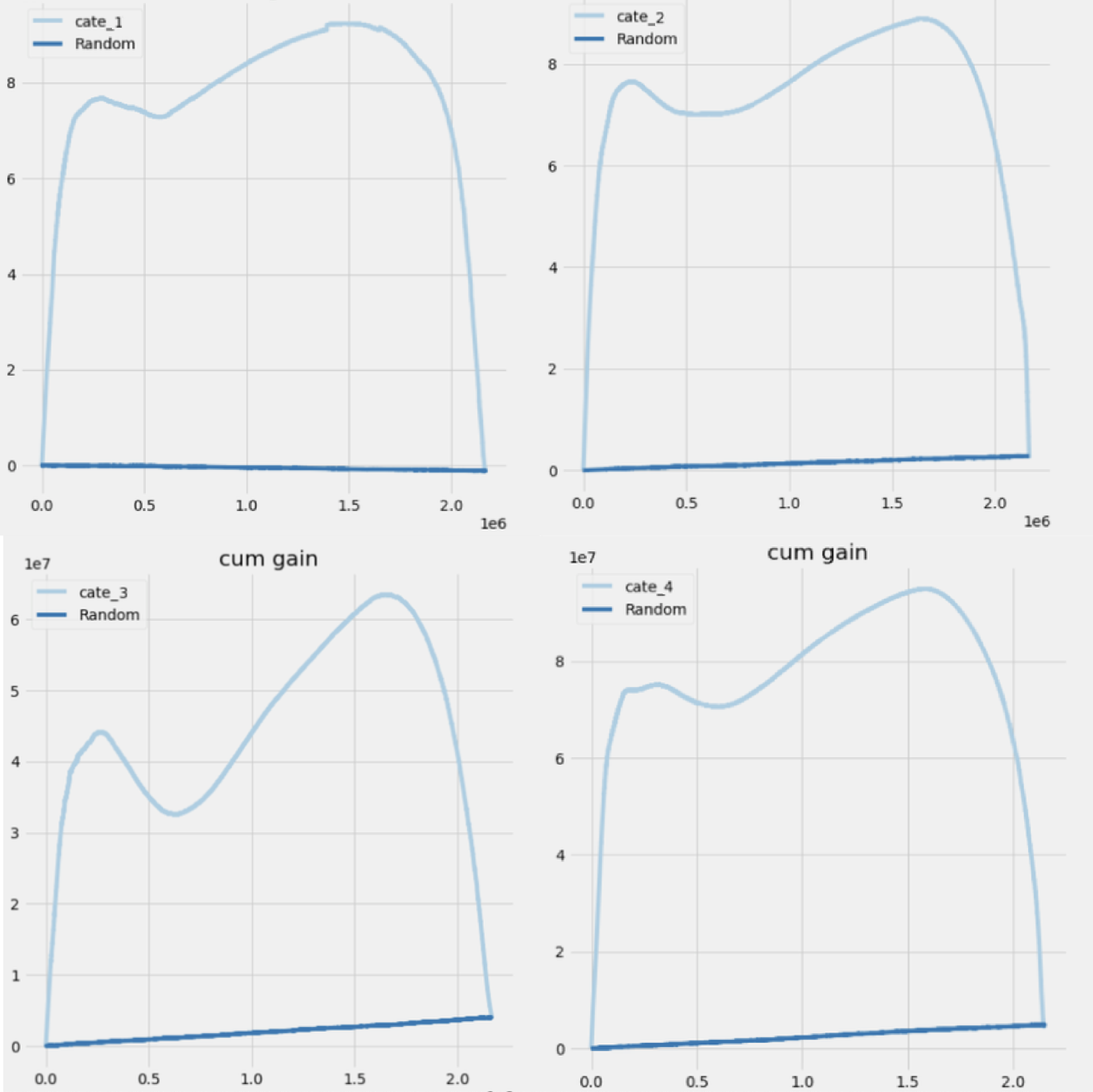}
    \caption{AUUC curves across treatments, the y-axis represents the cumulative incremental gains, and the x-axis is the proportion of the population targeted.}
    \label{fig:enter-label}
\end{figure}
\end{itemize}

\section{Conclusion and Future Work}
\subsection{Conclusion}
Our research highlights the intricate trade-offs between upselling intensity and user satisfaction. While increasing upselling intensity leads to immediate revenue gains, it also results in a decline in user recall and satisfaction. This effect was most pronounced in the TG1 group, where a bi-modal distribution of responses revealed polarized reactions to upselling strategies. These findings emphasize the need to carefully balance upselling intensity with user experience to maintain both engagement and revenue growth.

The causal uplift modelling framework allows us to tailor the intensity for each user, effectively managing these trade-offs and ensuring fairness in recommendations. Results from both the test and training datasets show that taking these models online holds significant potential, with an estimated revenue uplift of 10.7\%.

Further analysis, as shown in Figure [\ref{fig:sensitivity}], revealed that the guardrails mainly were impacted for new users. The offline CATE policy applied a 1x intensity for 80\% of new users, resulting in a 2.1\% improvement in conversion rates. While these offline results are promising, further validation is required in an online experimental setting to confirm the optimal trade-offs between revenue and user experience.

\subsection{Future Work}
Our future work will focus on testing the models online, exploring additional user features, and extending the experimentation to incentivise user behaviour upselling on the platform. Further, developing personalized upselling strategies that dynamically adjust intensity levels based on user behaviour and feedback could enhance revenue and user satisfaction.

\section{Appendix}
\subsection{Platform Implementation}
\begin{figure*}[htbp!]
  \centering
  \includegraphics[width=1\linewidth]{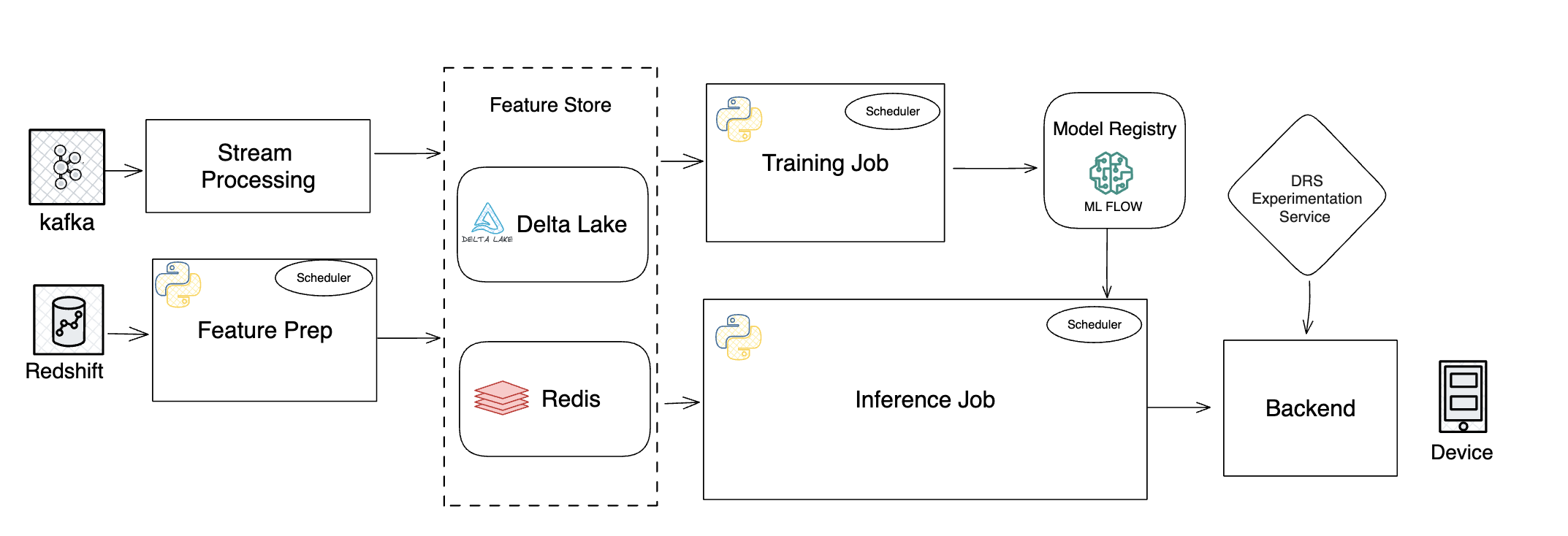}
  \caption{Personalised Upselling Model}
  \label{fig:PUM}
\end{figure*}

The "Deposit Amount Upselling" platform shown in figure \ref{fig:PUM} integrates several key components to predict deposit amounts and provide personalized upselling recommendations. Below is a summary of the implementation steps:

\begin{itemize}
    \item \textbf{Data Ingestion}: The platform ingests data from two primary sources: streaming events and backend data stored in the warehouse. 

    \item \textbf{Feature Aggregation/pre-processing}: We prepare the data by performing the following steps: label creation, gradient features calculation, outlier removal and feed it to the Deposit Amount Predictor model, ensuring the model leverages both real-time and historical data for accurate predictions.

    \item \textbf{Deposit Amount Predictor}: A machine learning model predicts the likely deposit amounts for users. The model outputs deposit amount classes (e.g., 0-20, 20-40, 1000-25000) and corresponding probability scores.

    \item \textbf{Upselling Recommendation}: The predicted amounts and scores are processed by an upselling policy, which employs an intensity layer to experiment using the model's predictions.
    
\end{itemize}

\subsection{Technical Stack}
\begin{itemize}
    \item \textbf{Data Ingestion}: Real-time data processing is performed using Kafka on our in-house stream processing platform \cite{streamverse2023}. Backend data is retrieved from SQL databases like Redshift.
    \item \textbf{Feature Store}: Processed features are stored efficiently in a feature store, supporting high throughput and low-latency retrieval. Delta Lake is used for offline storage, while Redis is employed for online storage.
    \item \textbf{Model Training and Prediction}: Causal Machine learning models are developed in Python using \cite{chen2020causalml} library. Airflow manages the scheduling for both batch and real-time inference scenarios.
    \item \textbf{Model Management}: ML Flow \cite{Zaharia2018AcceleratingTM} provides seamless model tracking and registry management.
    \item \textbf{Deployment and Inference}: We power and monitor our inference jobs on our in-house machine-learning platform, Darwin 
    \item \textbf{Backend Integration}: Recommendations are served into backend systems for delivery to end-user devices.
\end{itemize}

\bibliographystyle{ACM-Reference-Format}
\bibliography{main,blogs}

\end{document}